\pdfoutput=1
\documentclass{article}

\usepackage{graphicx}
\usepackage{hyperref} 
\usepackage{amsmath} 
\usepackage{booktabs}
\usepackage{url}
\usepackage{cleveref} 
\usepackage{float}
\usepackage{eurosym}
 \usepackage{natbib}

\makeatletter
\newcommand{\printfnsymbol}[1]{%
  \textsuperscript{\@fnsymbol{#1}}%
}
\makeatother

\title{Credit card fraud detection using machine learning: A survey}
\author{Dr. Yvan Lucas \thanks{equal contribution} \\INSA Lyon \\
	\and 
	Dr. Johannes Jurgovsky \printfnsymbol{1}  \\University of Passau}

\begin{document}

\maketitle

\begin{abstract}
Credit card fraud has emerged as major problem in the electronic payment sector. In this survey, we study data-driven credit card fraud detection particularities and  several machine learning methods to address each of its intricate challenges with the goal to identify fraudulent transactions that have been issued illegitimately on behalf of the rightful card owner.

In particular, we first characterize a typical credit card detection task: the dataset and its attributes, the metric choice along with some methods to handle such unbalanced datasets. These questions are the entry point of every credit card fraud detection problem.

Then we focus on dataset shift (sometimes called concept drift), which refers to the fact that the underlying distribution generating the dataset evolves over times: For example, card holders may change their buying habits over seasons and fraudsters may adapt their strategies. This phenomenon may hinder the usage of machine learning methods for real world datasets such as credit card transactions datasets. 

Afterwards we highlights different approaches used in order to capture the sequential properties of credit card transactions. These approaches range from feature engineering techniques (transactions aggregations for example) to proper sequence modeling methods such as recurrent neural networks (LSTM) or graphical models (hidden markov models).
\end{abstract}

\newpage
\tableofcontents

\bigskip


Both credit card fraud and its detection are very specialized domains that attract interest from a small highly specialized audience. However, we believe that with this survey we may help both researchers and practitioners who recently started working in this field to bootstrap the most elemental detection ideas and to get a general overview of the area.

Indeed, credit card fraud detection is a crucial asset for ensuring customer trust and saving money by preventing fraudulent loss. However, credit card fraud detection presents several characteristics that make it a challenging task.

\paragraph{Skewed class distributions} Most transactions are legitimate and only a very small fraction of the transactions has been issued by fraudsters. Some algorithms such as Support Vector Machines for example suffer a lot from class imbalance and can be completely unable to learn the minority class specific patterns (\citep{wu2003}, \citep{yan2003}). Solutions have been presented in the literature in order to either adapt the algorithms (cost based methods for example) to imbalance or solve the imbalance in the data (sampling based methods).
\paragraph{Dataset shift} Fraudsters tactics evolve over time. Some fraud strategies become out of date because of the constant cat and mouse play between fraudsters and experts, and new technologies show weaknesses that can be exploited by fraudster. Moreover, cardholders behaviours change over seasons. . The combination of the aforementioned factors makes typical algorithms become out of date or irrelevant. Real world algorithms have to be designed in order to be updated regularly or with ways to detect a decreasing efficiency. Besides, online learning schemes are very useful in order to prevent concept drift and covariate shift to affect the fraud detection systems. 
\paragraph{Feature engineering} 
Credit card transactions are represented as vectors of continuous, categorical and binary features that characterize the card-holder, the transaction and the terminal. However, most learning algorithms can't handle categorical feature without a preprocessing step in order to transform them into numerical features (\cite{kotsiantis2007}). Additionally, specific feature engineering strategies for credit card detection exist. These strategies either aim to make the features more significant for the classifier or to create additional features in order to enrich transaction data. 
\paragraph{Sequence modeling} A major issue of typical credit card transaction dataset is that the feature set describing a credit card transaction usually ignores detailed sequential information. Therefore, the predictive models only use raw transactional features, such as time, amount, merchant category, etc. \cite{bolton2001} showed the necessity to use features describing the history of the transaction. Indeed, credit card transactions of the same cardholder (or the same terminal) can be grouped together and compared to each other.
\paragraph{Interpretability} Aside from the aforementioned research questions, interpretability of the machine learning algorithm's decision is a crucial concern for the industry who is one of the targeted user group of applied data science research work such as credit card fraud detection. Research has been done in order to understand better on which heuristics model decisions are based (\cite{pastor2019} for example). This stake also holds for the understanding of performance metrics by non machine learning experts.

Each of these challenges have been tackled in various ways over the 20 past years.

In this survey, we highlight works from different teams that aimed to solve the different credit card fraud detection challenges in order to provide a sound starting point to anyone wanting to dig in the field of credit card fraud detection.

\section{Credit card fraud detection entry point: dataset and evaluation} 
\label{sec:intro}

\subsection{Credit card transactions dataset}

Payment data contains very sensitive private information about individuals and businesses and access to such data is highly restricted to only the data owners and the companies that manage the data. Which is the reason why there is no publicly available data set that would sufficiently reflect the magnitude and variety of real-world card payments and that could therefore be considered as a basis for studying the many interesting challenges present in this domain. 

The only credit card transactions we were able to find\footnote{Besides the confidential one used during our joint research project with Worldline} was a small sample of 280.000 transactions with PCA-transformed features\footnote{\url{https://www.kaggle.com/mlg-ulb/creditcardfraud}, Last access: 10.05.2019.} published alongside a study which was conducted jointly together with the exact same payment service provider~\citep{dal2015calibrating}. This scarcity of publicly available real-world data is problematic for the research community, since it compromises transparency, comparability and reproducibility of published findings. 

\begin{table} \centering
    \begin{tabular}{llp{0.35\linewidth}p{0.2\linewidth}}
    \toprule
    Category & Attribute & Description & Feature(s)\\
    \midrule
    Transaction & \underline{Time} & Date and time of the transaction  \vspace{0.3cm} & \texttt{weekday}, \texttt{hourOfDay}, \texttt{timestamp}\\
               		  & \underline{Type} & The kind of transaction issued. A real payment (e.g. e-commerce) or a virtual transaction (3D-secure ACS).  \vspace{0.3cm} & \texttt{ecom}, \texttt{3DSecureACS} \\
               		  & \underline{Authentication} & The kind of authentication experienced by the card holder: Chip-based PIN check, signature, 3D-secure. \vspace{0.3cm} & \texttt{emv}, \texttt{authentication}, \texttt{3DSecure} \\
               		  & Entry mode & The interaction modality between the card and the terminal: e.g. contactless, magentic stripe, manual entry.  \vspace{0.3cm} & \texttt{cardEntryMode} \\
               		  & Amount & The transacted monetary value.  \vspace{0.3cm} & \texttt{amount} \\
               		  & Fraud label & The manually assigned label indicating whether a transaction was a fraud. & \texttt{fraud}\\
    Card holder & \underline{Declarative profile} & Several attributes describing the card holder such as his country of residence, age or gender. \vspace{0.3cm} & \texttt{userCountry}, \texttt{age}, \texttt{gender}\\
    					  & \underline{Card} & Several attributes describing the credit card such as the credit limit, card type or expiration date. \vspace{0.3cm} & \texttt{creditLimit}, \texttt{cardType}, \texttt{expiryDate}\\
    Merchant & \underline{Declarative profile} & Two attributes describing the terminal: The merchant category code assigned to the terminal (e.g. shoe store, ATM, etc.) and the country the terminal is registered in (e.g. Belgium, Germany, etc.) \vspace{0.3cm} & \texttt{terminalCategory}, \texttt{terminalCountry}\\
     \bottomrule
    \end{tabular}
    \caption[Attributes of transactions]{Transactions are described along three groups of attributes: Transaction related, card holder related and merchant related. Underlined attributes are in fact collections of several attributes. Examples of attributes in the collection are listed in the description.}
    \label{tab:attributes}
\end{table}
	  
As listed in~\cref{tab:attributes}, transactions are characterized by several attributes. We can broadly categorize them into attributes characterizing the transaction itself, attributes characterizing the card holder/card that issues a transaction and attributes characterizing the merchant who receives the transaction. In addition, each card holder and each merchant is assigned a unique identifier. 

The fraud label is binary and it indicates the authenticity of a transaction. The label was assigned by human investigators after consultation with the card holder. The label takes the value \texttt{0} in most of the cases. Only when a transaction has been identified as having been issued by a third party and not the rightful card holder, is the label set to \texttt{1}, which flags the transaction as fraudulent. Throughout the paper we will refer to the label as the \textit{fraud label} or the \textit{class} of the transaction.

\subsection{Performance Measures} \label{sec:performance}

\subsubsection{Confusion matrix based evaluation}
\label{sec:confmat}
Traditional ways of evaluating machine learning classifiers use the confusion matrix describing the difference between the ground truth of the dataset and the model prediction.

\begin{figure}[h]
\centering
\includegraphics[width=1.0\linewidth]{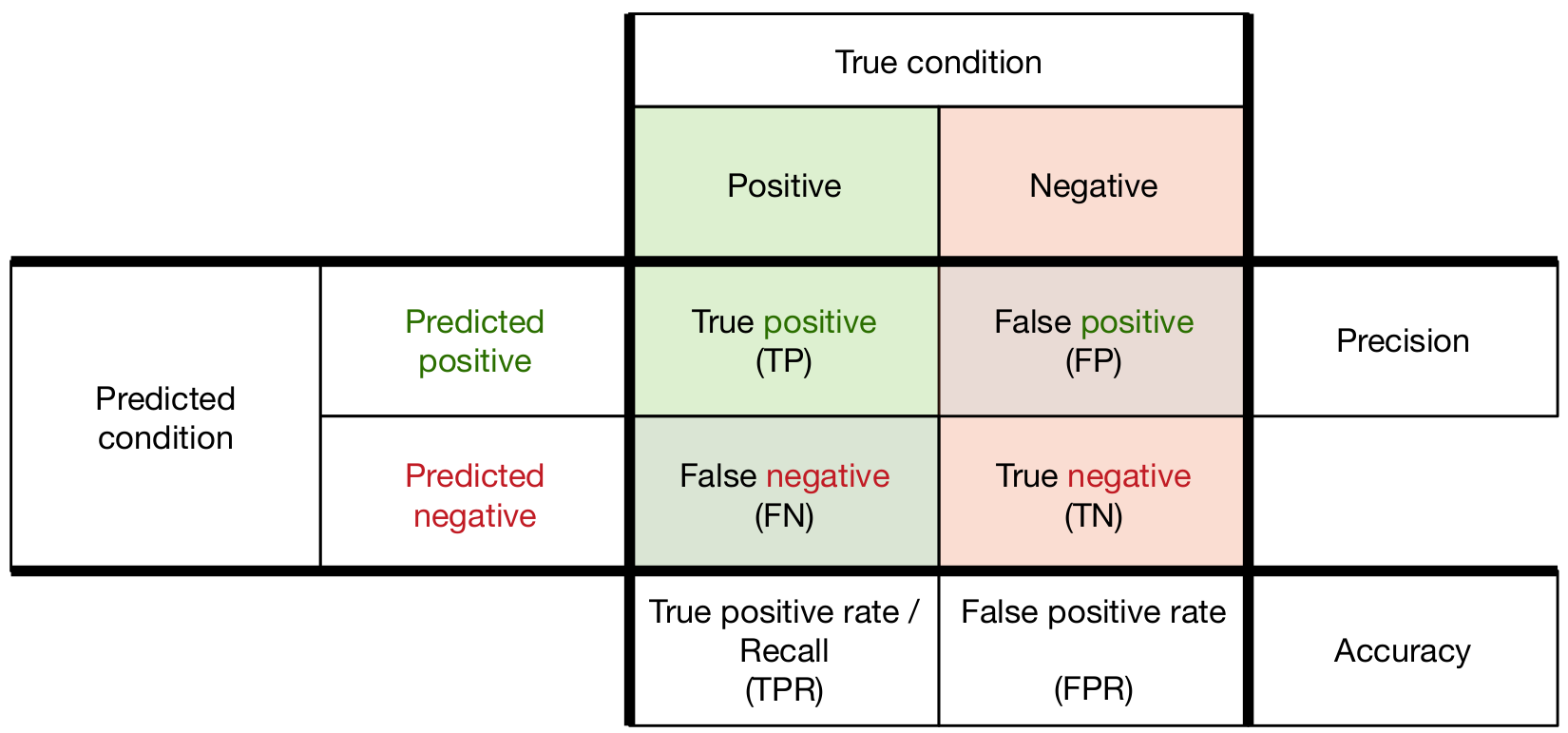}
\caption[Confusion matrix]{A confusion matrix displaying the four possible outcomes of a binary decision (colored boxes) and an excerpt of measures that quantify different properties of the decision maker in terms of ratios of these outcomes. Type 1 errors (false positives) and type 2 errors (false negatives) are shaded in gray.}
\label{fig:contigency-table}
\end{figure}

Using the confusion matrix, one can compute a variety of metrics in order to compare classifiers performances:
\begin{description}
\item[Precision ] Proportion of well classified examples among examples of the positive class: $$prec=\frac{TP}{TP+FP}$$
\item[Recall ] Also called true positive rate, proportion of examples of the positive class well-classified among all positive examples: $$rec=\frac{TP}{TP+FN}$$.
\item[Accuracy] Proportion of well classified examples among all the testing examples (similar to precision but for every class): $$acc=\frac{TP+TN}{TP+TN+FP+FN}$$
\item[F1 score] Precision and recall have opposite behavior: forcing a good precision is detrimental for the recall and vice versa. F1 score is the harmonic mean between precision and recall and therefore gives equal importance to precision and recall. For this reason, this metric which takes values between 0 and 1 is often considered a good by default metric for unbalanced classification tasks: $$F1=2*\frac{Precision * Recall}{Precision + Recall} = 2* \frac{2TP}{2TP+FP+FN}$$
\item[MCC] Matthews correlation coefficient is another score that is robust to imbalance \cite{david2011}. It is related to the chi-square statistic and takes values between -1 and 1. For binary classification: $$MCC=\frac{TP \times TN - FP \times FN}{\sqrt{(TP+FP)(TP+FN)(TN+FP)(TN+FN)}}$$
\end{description}

Confusion matrix based metrics are a good starting point for evaluating classifiers' performances. However these metrics are point estimators and some users may want to have a fine grained knowledge of how the classifiers are performing over all the domains of the testing set. Parametric evaluation of classifier performance allows for that more fine grained look at the classifiers' behaviours. 

\subsubsection{Parametric evaluation}
\label{sec:parametric}
Some machine learning classifiers like random forest, neural networks, SVM, logistic regression classifier, etc. output a set of $n$ predictions for each example corresponding to their confidence that each example belongs to each of the $n$ classes. 
\newline

By changing the value of a threshold parameter through all the possible values of probabilities obtainable with the machine learning classifier, one can monitor the efficiency of the prediction for each examples: from the most obvious to classify to the least obvious. In fine, parametric evaluation can be seen as an evaluation of all the possible confusion matrices obtainable given a prediction vector.
\newline 


When doing a parametric evaluation, two curves are usually plotted: the ROC (Receiving Operating Characteristic) curve and the precision-recall curve  
\begin{itemize}
\item The ROC curve's axes are: true positive rate (i.e. recall) for the y-axis and false positive rate for the x-axis. Both axes are proportion of the class they are representing: positive for the true positive rate and negative for the false positive rate. Therefore, the ROC curve is insensitive to unbalanced datasets: differences in class effectives won't affect the overall shape of the curve. 
\item The Precision-Recall curve axis are the precision for the y-axis and the recall for the x-axis. As for the axis of the ROC curve, the recall is a ratio of the positive class: it represents the proportion of example of the positive class identified. However, the precision is based on both positive class (TP) and negative class (FP) effectives. Therefore, a strong class imbalance (such as credit card fraud detection imbalance) would affect the precision values since the classification is harder. The precision recall curve is more punitive than ROC curve and consequently more challenging to improve: in a context of class imbalance, the precision value is closer to 0 and therefore the precision recall area under the curve decreases. Besides, we experienced that the precision recall curve is easier to read for industrial partners.
\end{itemize}

\begin{figure}[h]
\centering
\includegraphics[width=0.8\linewidth]{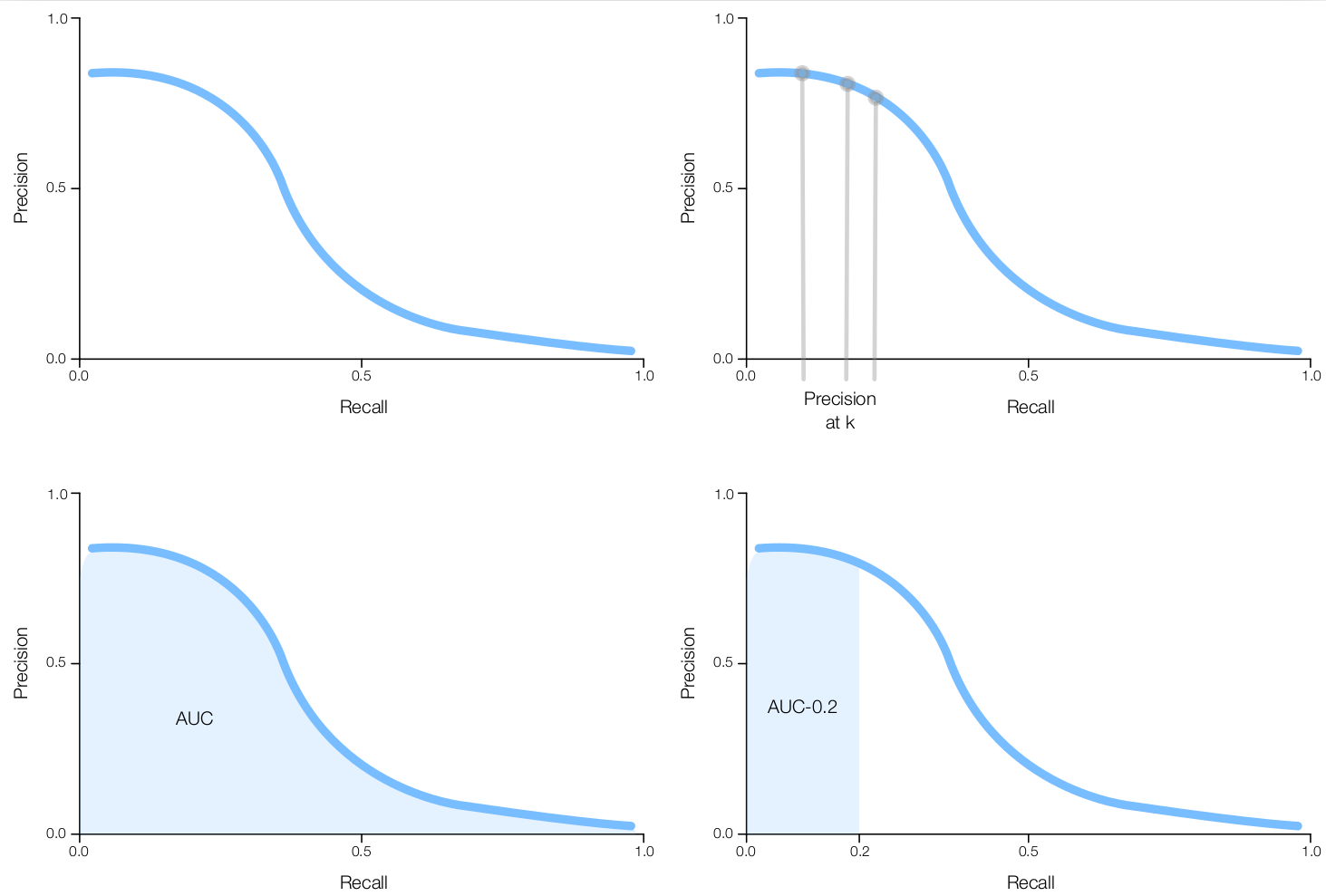}
\caption{Precision recall curve and area under the curve \textit{Top: Precision recall curve and precision at $K$, indicated by grey lines. Bottom: Area under the precision recall curve and area in the early retrieval range, calculated only over the interval $[0, 0.2]$.}}
\label{fig:aucpr}
\end{figure}

According to domain application needs, the area under a portion of the curve can also be calculated. For example, for credit card fraud detection evaluation, the need to calculate the AUC for values of recall below 0.2 was discussed since the experts can only investigate a limited number of alerts. Some others

Further details about the intricacies in interpreting ROC curves versus precision recall curves have been discussed in~ \citep{NoROCSaito2015} and~\citep{Davis2006}. Although the interpolation between points requires special care, the area under the precision recall curve (AUCPR) can serve as a single-valued aggregate measure of a classifier's performance~\citep{Hand2009}. 
In~\cref{fig:aucpr} we illustrate several measures that can be derived from a precision recall curve:
\begin{itemize}
\item Precision at $K$ (P@K): The proportion of true-positives in the set of $K$ transactions with the highest scores. Each choice of $K$ corresponds to a specific level of recall on the x-axis. 
\item Area under the precision recall curve (AUCPR): The integral of precision over all levels of recall over the interval $[0, 1]$.
\item Area in the early retrieval range (AUCPR@0.2): The integral of precision over recall levels from $0$ to $0.2$. Instead of reporting precision at several values of $K$, we use this early retrieval range to reflect the application-specific focus on high precision in the range of low recall.
\end{itemize}

\section{Learning to classify unbalanced datasets} \label{sec:imbalance}
Typical fraud detection tasks present a strong imbalance in the representations of the classes: Usually there is a majority genuine class that outweighs the minority fraudulent class \cite{Ali2019}.
\newline

Most learning algorithms are not adapted to a strong imbalance in the representations of classes in the dataset. Therefore, learning from unbalanced dataset is a challenging research question that needs to be considered (\cite{batista2000}). \cite{weiss2001} and \cite{estabrooks2004} observed that an unbalanced dataset leads to a decrease in the performance of standard learning algorithms such as SVM or random forest. In fact, class imbalance have been shown to produce a negative effect on the performance of a wide variety of classifiers (\cite{chawla2005}). \cite{japkowicz2002} have shown that skewed class distributions affects negatively the performance of decision trees and neural networks. \cite{visa2005} drew the same conclusion for neural networks. \cite{kubat1997}, \cite{mani2003} and \cite{batista2004} showed the weakness of k Nearest Neighbors algorithms to imbalanced datasets. \cite{yan2003} and \cite{wu2003} showed that SVM were also weak to skewed class distributions. Overall, most of the classifiers suffer from class imbalance, some more than others.
\newline

The strategies used to tackle this issue operate at two different levels: some aim to solve the imbalance of the dataset beforehand whereas others want to adapt the learning algorithm to make it robust to an unbalanced classification task (\cite{chawla2004}, \cite{pozzolo2015}).
\newline

The data level strategies are called like that since they happen during the preprocessing time, before any learning algorithm is applied. They consist in rebalancing the dataset in order to compensate the structural imbalance. The algorithmic level strategies involve methods where the algorithms are designed especially to deal with an unbalanced task and methods where the classification costs of the algorithm are adapted in order to reflect the priorities of the unbalanced classification task. The latter are called cost sensitive classifiers (\cite{elkan2001}).
\newline

The sampling based solutions to handle imbalance will be described in section \ref{sec:sampling_desc}. The model based methods will be briefly highlighted in section \ref{sec:algo_desc}

\subsection{Sampling methods}
\label{sec:sampling_desc}
\begin{figure}[t]
\centering
\includegraphics[width=\textwidth]{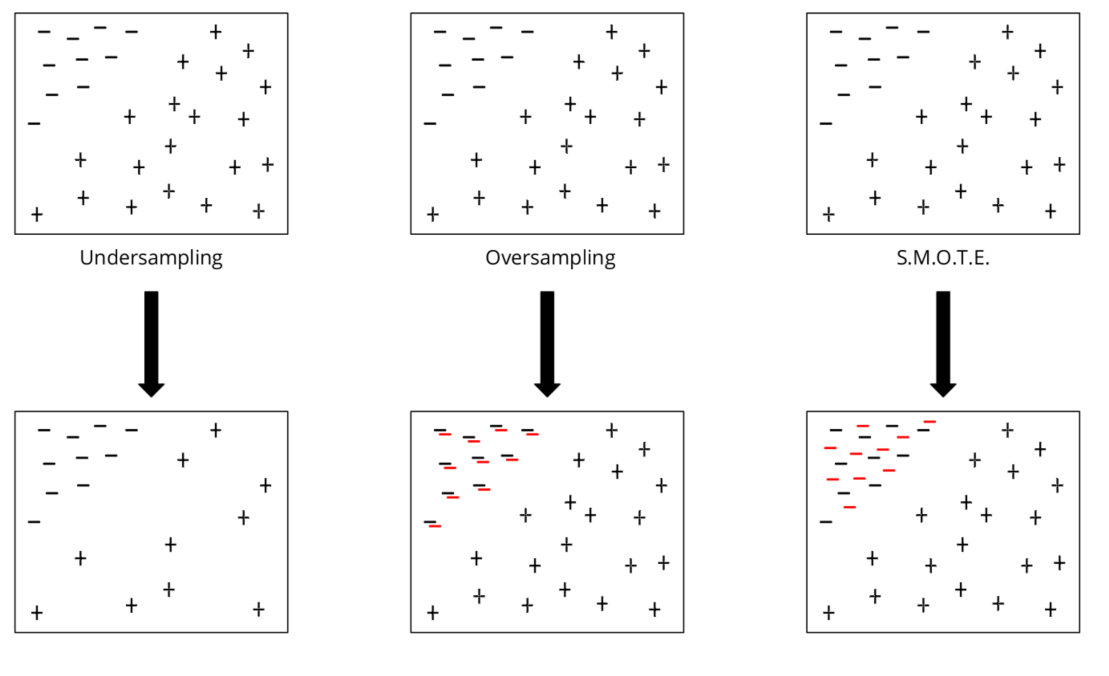}

\caption[Sampling strategy for unbalanced classification.]{Sampling strategy for unbalanced classification. \textit{(The $+$ symbols refer to the majority class whereas the $-$ symbols refer to the minority class. The symbols colored in red are newly created symbols by the oversampling strategy or by the SMOTE strategy(\cite{chawla2002}))}}
\label{fig:SMOTE}
\end{figure}

Sampling methods consist in reducing the class imbalance by changing the class effectives in the dataset. In this section we will first present two naive methods that reduce class imbalance in the dataset by adjusting class ratios: undersampling and oversampling. Afterwards we will introduce the Synthetic Minority Oversampling Technique strategy (SMOTE) that aims to create new examples of the minority class (\cite{chawla2002}). The different strategies are illustrated in figure \ref{fig:SMOTE}.
\newline

Undersampling consists in rebalancing the dataset by taking less examples from the majority class in order to match the absolute number of examples of the minority class (\cite{drummond2003}). The assumption made with undersampling is that the majority class contains redundant information that can be removed. The main issue with undersampling is that, when the imbalance is too pronounced, too many examples from the majority class need to be taken away. This may cause a decrease in the performance of the algorithm due to a lack of data.
It is worth mentioning that undersampling speeds up the learning phase, which makes it an interesting choice when in presence of an unbalanced dataset.
\newline

Oversampling consists in rebalancing the dataset by repeating random examples of the minority class (\cite{drummond2003}). The main issue of oversampling is that repeating some examples doesn't add information and may lead to an overfitting of the learning algorithms. Besides, oversampling increases the learning time since it makes the train-set artificially bigger.
\newline

In 2002, \cite{chawla2002} introduce SMOTE, a strategy to reduce class imbalance by creating new examples in the neighbourhood of the minority class. The creation of new minority class example is done by averaging neighboring examples of the minority class. For categorical features, the averaging consists in a majority voting for the category: the category mostly represented in the neighbourhood becomes the category of the corresponding categorical feature of the newly created example. They show an increase in the performance of classifier when using this rebalancing strategy. However SMOTE does not consider the label of neighboring examples when creating examples from the minority class. This can lead to an overlap between minority and majority classes. ADASYN \cite{he2008} and Borderline-SMOTE \cite{han2005} tackle this issue by taking into account only neighboring examples from the minority class.

\subsection{Model based methods}
\label{sec:algo_desc}

Some solutions were published in order to strengthen ensemble based methods when in presence of class imbalance. EasyEnsemble (\cite{liu2009}), UnderBagging (\cite{wang2009}) and SMOTEboost (\cite{chawla2003}) are ensemble based method that are adapted by design to skewed class distributions. SMOTEboost (\cite{chawla2003}) combines boosting with SMOTE in order to provide balanced datasets to each boosting weak learner. Similarly, EasyEnsemble (\cite{liu2009}) uses undersampling in order to create different balanced datasets for the boosting learners. In UnderBagging (\cite{wang2009}) undersampled datasets are used to train several learners whose predictions are aggregated with bagging.
\newline

An other leverage for dealing with class imbalance is to adapt the metrics used to evaluate the models and the costs used to train some of them in order to counterbalance class imbalance. Indeed, some real world imbalanced classification tasks such as credit card fraud detection give more importance to correctly identify the element of the minority class than those of the majority class. In the case of cost sensitive learners, the misclassification cost can be adapted in order to reflect this priority. \cite{shirazi2010} present cost sensitive SVMs. \cite{kukar1998} introduced earlier the possibility of cost adaptation for neural networks. Adapting the misclassification cost for imbalanced problem has also been studied for boosting methods (\cite{sun2007}, \cite{wei1999}).

\section{Dataset shift detection and adaptation} \label{sec:drift}

Many machine learning tasks make the assumption that the random variables in the data are i.i.d. (independent and identically distributed) among the train and the test sets. That property ensures that the decision function learned by the classifier on the train set is valid on the testing set. However in some case the i.i.d. hypothesis isn't valid: there are changes in the data distribution between the training and testing set. These changes of the data distribution can decrease drastically the performances of the classifiers (\cite{pozzolo2014}, \cite{abdallah2016}, \cite{rodriguez2008}) and need to be taken into account.
\newline

The term "dataset shift" is used in order to include the phenomenons where the i.i.d. hypothesis isn't valid. It is defined by \cite{torres2012} as "cases where the joint distribution of inputs and outputs differs between training and test stage"
\newline

Dataset shift of various types have been shown to affect classifiers performance. \cite{rodriguez2008} showed that complex classifiers are more robust to population drift than simple classifiers. 



\subsection{Different types of dataset shift}
Dataset shift is a broad word that refers to a change in the joint distribution of the target variable and the input features $p(x,y)$. However, since $p(x,y)\quad = \quad p(y|x)*p(x)$, the causes of the change in the joint distribution can be different. Several shifting phenomenon can be identified (\cite{torres2012}, \cite{storkey2009}, \cite{gao2007}, \cite{shimodaira2000}):
\begin{itemize}
\item \textit{Covariate shift} was formerly called \textit{population drift}. It refers to the phenomenon when the conditional distribution of the classes with respect to the training data doesn't change from the train set to the test set, but the distribution of the data do. For example in the credit card fraud detection settings, the phenomenon where the buying behaviors change over seasons but the fraudulent strategies don't can be seen as \textit{covariate shift}. It is formally defined as:
\begin{equation}
p_{train}(y|x) = p_{test}(y|x) \quad \textrm{and} \quad p_{train}(x) \ne p_{test}(x) 
\end{equation}
\item \textit{Prior probability shift} is an other subtype of dataset shift. It is also sometimes called \textit{shifting priors}. \textit{Prior probability shift} refers to a change in the class distributions: for example when the proportion of fraudulent transactions in the training set is way smaller than for the testing set. It may cause a over-representation or an under-representation of one class in the prediction of machine learning classifiers. It is formally defined as:
\begin{equation}
p_{train}(y|x) = p_{test}(y|x) \quad \textrm{and} \quad p_{train}(y) \ne p_{test}(y) 
\end{equation}
\item  \textit{Concept drift} is the type of shift that appears the most in the credit card fraud detection literature.\textit{Concept shift} or more often \textit{concept drift} refers to the phenomenon where the conditional distribution of the target classes with respect to the input variables changes. For example when fraudster adapt their strategies and the classifier is not able to detect fraudulent transactions efficiently anymore. It is formally defined as:
\begin{equation}
p_{train}(y|x) \ne p_{test}(y|x) \quad \textrm{and} \quad p_{train}(x) = p_{test}(x) 
\end{equation}  
\end{itemize}

\subsection{Detecting dataset shift}

Most of the studies passively detect dataset shift when a decrease in classifiers performance on the test set is witnessed(\cite{pozzolo2015}, \cite{gomes2017},...). In order to prevent the decrease of the classifier performance, some teams aimed to estimate more realistically the expected performance of classifiers in presence of dataset shift. For example, \cite{webb2005} pointed out the weaknesses of the ROC evaluation in presence of \textit{prior probability shift}. Earlier, \cite{kelly1999} stated that the weaknesses of classifiers in presence of covariate shift and prior probability shift can be highlighted with the decrease of classifiers performance for their loan risk prediction task. Additionally, \cite{sugiyama2007} discussed the issues appearing when classifier performance expectation is set using a cross validation hyper-parameter optimization: cross validation expectations can't be trusted if the distribution changes between train and test set. They proposed a cost based approach by comparing the data distribution $p(x)$ in the train and test set and weighting more the data points of the train set that are similar to those of the test set for the cross validation optimization. If the data point is typical of the train and test distribution, then it is weighted more for the evaluation of the model. 
\newline

\cite{barddal2019} provided interesting results where they showed how to describe in advance the concept shift in two dataset (Airlines and Sea \footnote{\url{https://moa.cms.waikato.ac.nz/datasets/}}) by looking directly at the changes in the decision function through the changes of feature importance in an adaptive random forest classifier (described in next paragraph) \cite{gomes2017}. They could therefore predict the direction of the dataset shift and the resulting change in the conditional distribution $p(y|x)$.  Earlier, \cite{lane1998} tried to quantify the direction of drift on a continuous one dimensional axis by looking at which examples get misclassified when concept drift happens. An optimization of the threshold value of their classifier was managed in the decision process of their user identification for computer security task. Similarly, \cite{wang2003} wanted to precisely characterize concept shift by comparing the class prediction between two classifiers trained before and after a concept shift (old training data vs recent training data). The examples that present a conflict for class prediction can then be studied in order to understand which changes happen between the conditional distributions $p_{old}(y|x)$ and $p_{new}(y|x)$
\newline

\begin{figure}[H]
\centering
\includegraphics[width=0.95\textwidth]{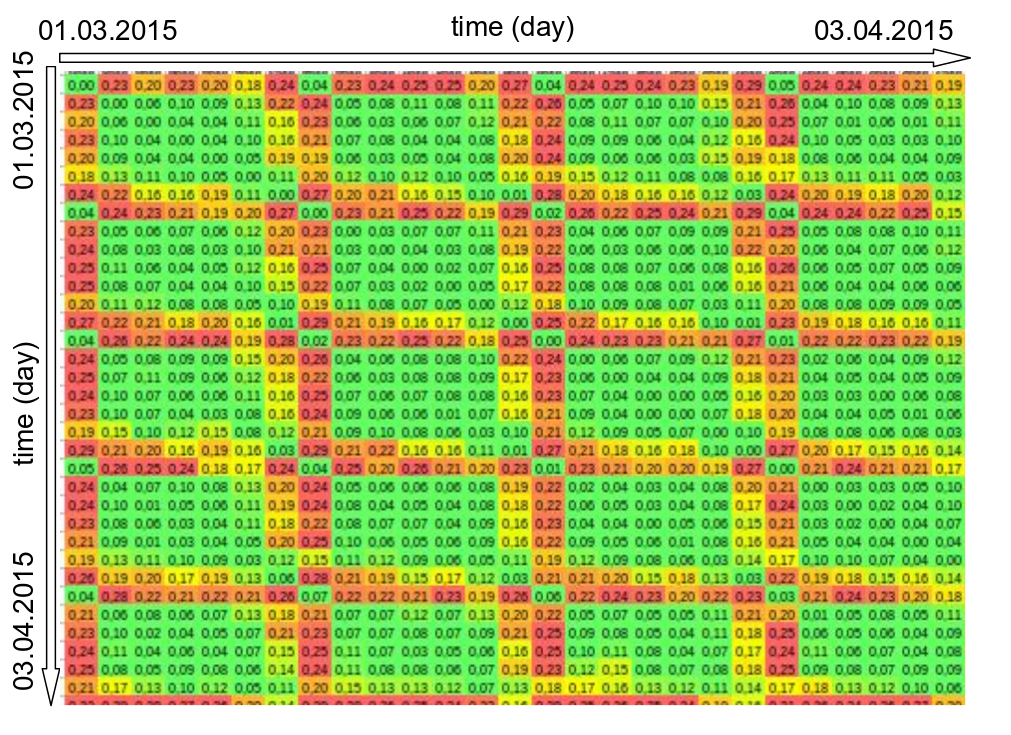}
\caption[Zoomed distance matrix (covariate shift) for the face-to-face transactions.]{Weekly periodicity observed in the dataset shift for face-to-face transactions. \textit{(centile based coloration: green $\Leftrightarrow$ similar days, red $\Leftrightarrow$ dataset shift} \cite{Lucas2020}}
\label{fig:lucas2019}
\end{figure}

\citep{lucas2019} introduced a technique to detect and quantify the extent of covariate shift between credit card transactions from different days. For each pair of days $(a, b)$:
\begin{itemize}
\item they collect all transactions from these days $D = D_{a} \cup D_{b}$ and label the transactions according to the day they came from.
\item They partition the sets of transactions and train a classifier to distinguish the day from which each transaction comes.
\item They evaluate the classifier performance and use this value as a distance measure between days: the better the performance, the easier the differentiation of the days, the more different the transactions of these days, and vice versa. 
\end{itemize}  
Through manual inspection and after clustering, the authors can show that low ”day distinguishability performance” clearly corresponds to calendar clusters of weekdays, Sundays and work holidays or Fridays and Saturdays (see figure \ref{fig:lucas2019}). 

\subsection{Strategies in presence of dataset shift}

\paragraph{Data based methods}
Some authors chose to handle dataset shift by adapting the training and testing set in order to erase the shift between these sets.

\cite{gretton2009} used reproducing kernel Hilbert spaces (RKHS) in order to weight instances in the training set so that the averages of the training set features correspond to those of the test set features. This kernel mean matching approach is mostly used in the context of covariate shift since it allows to make the distribution of the data in the train and test set more similar. \cite{bickel2007} leveraged this kernel mean matching approach for data streams. They derived the kernel mean matching approach in order to integrate it in the logistic regression classifier used and to update it over time. Similarly to the kernel mean matching approach, \cite{torres2013} proposed a method in order to achieve transfer learning between different cancer diagnosis datasets. They used genetic programming in order to create a regressor that transforms the features of one dataset to the features of an other dataset in order to be able to apply a pretrained classifier without adapting it.
\newline

In addition to kernel mean matching techniques, \cite{klinkenberg2004} propose to train SVMs with an adaptive time window in order to prevent covariate shift.

\paragraph{Classifier based methods}
An other way to prevent concept drift is to update regularly the models in order to adapt them to the dataset shift.
\newline

At first, sliding windows based approaches have been explored. The idea is that by retraining them regularly using fresh data, the classifiers will stay up to date with dataset shift events. \cite{widmer1996} used a sliding window strategy for their FLORA4 framework: they trained regularly new classifiers on recent and trusted data but stored the bad performing ones for hypothetical later usage. \cite{lichtenwalter2009} also leveraged the sliding window strategy for imbalanced data streams with concept drift: they re-trained regularly C4,5 decision trees classifiers in order to prevent covariate shift. Moreover, they adapted the information gain calculation used when building decision trees by incorporating the Hellinger distance  in order to increase the robustness of the C4,5 trees to class imbalance and to prior probability shift. The Hellinger distance is a symmetric and non-negative measure of similarity between two probability distributions. The equation \ref{fig:hellingerdist} corresponds to the Hellinger distance between two discrete probability distributions with equal effectives $P=(p_{1}, ..., p_{k})$ and $Q=(q_{1},..., q_{k})$.

\begin{figure}[H]
\begin{equation}
d_{H}(P, Q) = \sqrt{\sum \limits_{i \in V} (\sqrt{P})-\sqrt{Q})^2} 
\end{equation}
\caption{Hellinger distance: Measure of distributional divergence between two probability distributions P and Q}
\label{fig:hellingerdist}
\end{figure}

\begin{figure}[H]
\centering
\includegraphics[scale=0.35]{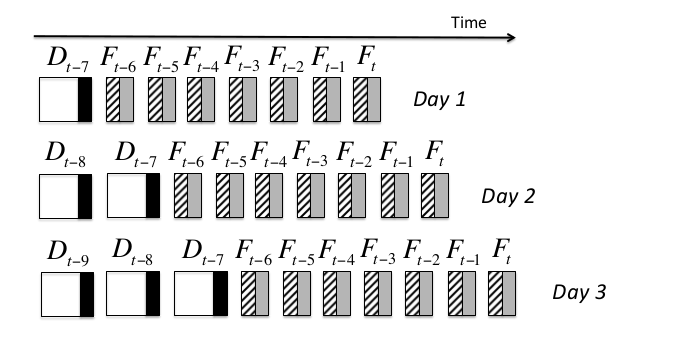}
\caption{Aggregating ground truth examples with investigators feedbacks (\cite{pozzolo2015})}
\label{fig:window}
\end{figure}

\cite{pozzolo2015} proposed a sliding window method for data streams with delayed information on the same credit card transactions dataset used in this thesis (see figure \ref{fig:window}). Since the ground truth for the class of credit card transactions is delayed by a week in average, the labels of the most recent transaction aren't known. They proposed to aggregate the predictions of a classifier on the old ground truth example with the predictions of a classifier trained on the feedbacks of the investigators on alerts raised previously by the fraud detection system. They showed that aggregating predictions provided better fraud detection than pooling all the examples of the ground truth and investigators feedback together.

Instead of preventing dataset shift by constantly re-training the classifiers, one may want to retrain them only if they prove to be lackluster. \cite{kolter2003} proposed to train again the classifiers when their performance drop. In order to prevent missclassification for data streams, \cite{hoens2011} used Hellinger distance on all the features of the dataset in order to detect early a covariate shift and which features are affected by it. They would therefore be able to know which trees of a random forest classifier will be affected by the dataset shift and need to be re-trained.
\newline

More recently, authors in \cite{gomes2017}, \cite{barddal2019bis} proposed an autonomous architecture for random forest classifier introducing a variant to Hoeffding Adaptive Tree in order to include a drift detector inside decision nodes to monitor the internal error rates of the tree, thus, continuously selecting features and building its predictive model over time. The Hoeffding Adaptive Trees (also called Very Fast Decision Trees) leverage the observation that using only a subset of examples in order to build a decision tree is sufficient to obtain good prediction \cite{bifet2009}. This computationally efficient and memory efficient tree architecture is therefore an asset in dataset shift contexts where models or parts of the models have to be retrained regularly.   

\section{Feature Engineering} \label{sec:features}

\subsection{Feature Encoding}
Most learning algorithms can not handle categorical variables with a large number of levels directly. Both supervised~\citep{kotsiantis2007} and unsupervised methods~\citep{lam2015clustering} require preprocessing steps to transform them into numerical features. In this section, we discuss several common solutions for encoding categorical attributes such that they can be used in learning algorithms.

\paragraph{Label Encoding and One-hot encoding} The most direct encoding of categorical attributes with $K$ different values is to create discrete variables $x \in \{1, 2, \dots, K\}$. Depending on the learning algorithm, label encoding might not be adequate to represent categorical attributes. The distance on integers typically does not reflect the distance on the attribute's values and would cause misleading results in distance-based algorithms. On the other hand, One-Hot encoding is particularly convenient and commonly used in machine learning literature~\citep{bishop2006pattern}. A categorical variable is represented by a $K$-dimensional vector $\mathbf{x}$, in which one element $x_k$ is set to $1$ and all remaining elements are set to $0$. 
\paragraph{Likelihood Encoding} (also known as target encoding) The value of a nominal attribute is encoded by its relative frequency of showing up in the positive class in a training set. If the $k$-th state of a nominal attribute appears $n_p$ times in the positive class and $n -n_p$ times in the negative class, the state gets assigned the value $x_k = \frac{n_p}{n - n_p}$. This type of encoding requires class labels and it assumes that the nominal values' representation in both classes provides sufficient information to distinguish the values. Therefore, we consider this technique rather as a feature engineering method than an actual encoding technique.
\paragraph{Vector Embeddings}
The values of categorical attributes may not always be orthogonal as suggested by a one-hot coding scheme but they rather exhibit some notion of similarity relevant to the prediction task. For instance, an attribute like the terminal's country can be endowed with different notions of similarity based on geographical locations, economic alliances or cultural backgrounds. It is common practice in machine learning to account for such expert knowledge through purposefully encoded attributes. Even if the main motivation is not an integration of a-priori knowledge, categorical attributes with many different levels partition the set of examples into many small sets for which it might be challenging to obtain reliable parameter estimates. Vector embeddings are the outcome of unsupervised encoding techniques with the aim to embed the values of a categorical variable in a continuous valued vector space. The role of unsupervised methods is thereby to estimate a map $c: X \to V^d$ from categorical values $X$ to points in a $d$-dimensional vector space $V^d$ such that distances between points reflect some sought notion of similarity between categorical values. We keep this superficial description of vector embeddings because there is an abundance of techniques by which such maps can be created, each with its own definition and interpretation of $c$ and $V$. For instance, the rows and columns in a tf-idf - matrix~\citep{JONES1973619} in information retrieval, the projections on the first $d$ principal components in dimensionality reduction, the mixture coefficients in probabilistic models with mixture distributions~\citep{blei2003latent} or the weights in neural networks~\citep{mikolov2011}. Similar to the likelihood encoding, the continuity in the embedding space is assumed to permit reasonable interpolation between otherwise distinct categorical values. \cite{russac2018} for example, successfully used Word2Vec embeddings in order to replace one-hot encoding for a distance based classifier (logistic regression) in a credit card fraud detection task.

\subsection{Feature Engineering}

In credit-card fraud detection, a fraud is supposedly not exclusively a property of the transaction itself but rather a property of both the transaction and the context it appeared in. While a transaction might arouse suspicion if, for instance, the spent monetary amount is very high and it takes place at a particular type of merchant at a certain time of day, such fraud detection strategy considers transactions in isolation from each other. Neither the historic purchases of the customer nor other transactions from the merchant he interacted with are taken into consideration. Two transactions occurring in geographically distant locations at almost the same time are often considered as suspicious patterns ("collisions"). Also, it may be that the volume of transactions per time (in an account) is more indicative of fraud than the absolute attributes of any transaction~\citep{whitrow2008}. In order to go beyond transaction-level detection, it is crucial to choose a suitable transaction context and extract relevant features from it.  

In general, most studies seem to agree that the Recency - Frequency - Monetary Value (RFM) framework provides the necessary information for discovering frauds. Recency refers to the time since the last transaction, frequency captures the volume of transactions per time interval and the monetary value corresponds to the total money spent within the interval. Some studies consider only frequency and monetary value \citep{krivko2010hybrid}, whereas others also integrate the time difference (\citep{vlasselaer2015} \citep{bahnsen2016}). From this perspective, aggregations are convenient because they allow to condition the RFM statistics on transaction types. It should be noted that there is no consensus on how to choose the most suitable time intervals. Usually, researchers apply reasonable rules of thumb and compute aggregates over days, weeks or months.

\paragraph{Rules}

\paragraph{Time differences} Since frauds tend to appear in bursts of several transactions within a short time window, the time difference between consecutive transactions seems to be a promising indicator of fraud. For any two consecutive transactions $\mathbf{x}_{t-1}$ and $\mathbf{x}_t$ with their time stamps denoted by $\mathbf{x}_{t-1}^{(\operatorname{Time})}$ and $\mathbf{x}_{t}^{(\operatorname{Time})}$ the \texttt{tdelta} feature of the $t$-th transaction is:
\begin{equation} \label{eq:tdelta}
\operatorname{tdelta}_t = \mathbf{x}_{t}^{(\operatorname{Time})} - \mathbf{x}_{t-1}^{(\operatorname{Time})}
\end{equation}
These time differences feature have been shown to provide relevant insight to classifiers in a credit card fraud detection problem (\citep{jurgovsky2018}).

\paragraph{Feature aggregations} Most feature engineering work in credit card fraud detection follows transaction aggregation strategies such as the ones described by \citep{whitrow2008}. The idea is to characterize the spending pattern in an account, the so called \emph{activity record}, by accumulating transaction information over time. Through exploratory data analysis, the authors identified several salient aspects of transactions as being relevant to fraud detection, such as the number of recently issued transactions and the total monetary value of these transactions. It is important to note that aggregates do not require any label information and, therefore, are entirely unsupervised. While aggregates function as account-level features, they can be turned into transaction-level features by updating the activity record with each new transaction. Most commonly, the statistics from the activity record are simply added as supplementary features to the basic feature set when training a fraud classifier. This aggregation strategy was found to work well in commercial applications and it is also an integral part of research in the academic field with broad usage in a number of studies (\citep{bhattacharyya2011},\citep{jha2012},\citep{bahnsen2013},\citep{sahin2013cost},\citep{pozzolo2014}). 

Since then, the proposed aggregation strategy served as generic template for developing new features. For instance, \citep{jha2012} and \citep{bahnsen2016} do not aggregate over specific transaction types but instead on those transactions whose attribute values are identical to the current transaction. This difference is subtle but it has severe implications for the semantics of the aggregated value. For example, \citep{whitrow2008} calculate the number of transactions issued in the last 24 hours at a specific merchant (e.g. "MyFancyShoeStore") for all transactions in the dataset. The value of this aggregate is probably zero for most transactions of most users but its semantic is consistent across all transactions. The aggregates defined in \citep{jha2012}\citep{bahnsen2016} calculate the number of transactions issued in the last 24 hours at the \emph{same} merchant as the current transaction. Finally, feature aggregates can be easily integrated in any existing classification process and they seem to readily improve the prediction accuracy. 

\paragraph{Trading entropy}  \citep{Fu2016} propose \emph{trading entropy}, a feature that is supposed to quantify the amount of information added to a user's purchase history when he issues a new transaction. As an example, they define the feature based on the amount spent and the merchant. In a user's history, they consider all but the most recent transaction and calculate how the total amount is distributed over the merchants. They quantify the uncertainty in this distribution by means of the Shannon entropy. The trading entropy is then the difference between the entropy of this distribution and the distribution they obtain after adding the most recent transaction to his history. 

\paragraph{Anomaly scores} Another interesting approach for extracting expressive features is discussed in \citep{vlasselaer2015}. Apart from the detection system the authors maintain a bipartite graph in which nodes correspond to merchants or card holders and edges correspond to transactions between these entities. The edge weight is determined by the volume of transactions exchanged between a card holder and a merchant, and it decays exponentially with time. With a page-rank style algorithm the authors propagate the fraud label of a transaction to directly connected or transitively reachable card holder and merchant nodes. The authors extract network features from the graph which measure the current exposure of each node to fraud. These features include a score for the card holder, the merchant and the transaction, aggregated over short, medium and long time periods.

\paragraph{External data integration} \citep{wetice2017} explored external data sources for augmenting transactions with information beyond the raw attributes. They extracted indicators of public holidays from calendars and used them as additional features of transactions. They also extracted embeddings from the DBpedia knowledge graph for encoding the originally categorical country tokens as points in a continuous vector space. Their results suggest a better generalization of the trained classifier and they hypothesize that this improvement results from the continuum between semantically related countries. Later, \citep{jurgovskythesis} supported this hypothesis by encoding country tokens in terms of several demographic statistics, such as a country's population, its gross domestic product or the corruption index. This explicit encoding led to a comparable performance increase as the unsupervised country embedding approach. Overall, encoding categorical tokens with semantic embeddings that can be extracted in an unsupervised fashion from public knowledge bases appears promising because such an approach does not require any manual effort and it readily improves the detection performance.

\section{Sequence modeling for fraud detection} \label{sec:sequence}

Sequence modeling is a major machine learning research question. In this section we will split the field with respect to three main approaches: the similarity-based approach that aims to learn the joint distribution of the events in a sequence in order to compute a similarity score, the model-based approach that aims to find the labels of the events in a sequence using data-driven strategies and the counting-based approach that aims to assess a likelihood score by counting the number of occurrences in a reference dataset.

\subsection{Similarity-based approaches}
\subsubsection{Graphical models}
Hidden Markov models (HMMs) are the most used model in the field of anomaly detection for discrete sequences according to \cite{chandola2012}.

They comprise two matrices (see figure \ref{fig:HMM1}):
\begin{itemize}
\item The \textit{transition matrix} describes the succession of the hidden states. It reflects a multinomial distribution from all hidden states at time $t$ to all hidden states at time $t+1$. The hidden states obey to the Markov property. Indeed, the hidden state at the time $t+1$ only depends on the hidden state at the time $t$: There is no memory in the distribution of successive hidden states, only conditional probability given the immediate previous hidden state. Formally, it means that conditional distributions between hidden states $t$ in a sequence of size $n$ can be simplified as following:
$$p(t_{k}|t_{k-1},t_{k-2},..,t_{1},w_{k-1},w_{k-2},w_{1}) = p(t_{k}|t_{k-1}) \;  k\in\{1,..,n\}$$
\item The \textit{emission matrix} describes the conditional distribution of the observed variable $w_i$ given the hidden states $t_i$. The probability distribution changes with respect to the observed variable properties. For example it can be considered multinomial for categorical observed variables or gaussian for continuous observed variables.
\end{itemize}


\cite{srivastava2008} used multinomial HMMs in the context of credit card fraud detection. They modeled the sequence of amount of users as a categorical feature with 3 values: "low amount" for the transactions with amounts less than 50 \euro{}, "mid amount" for transactions with amounts between 50 and 200 \euro{} and "big amount" for transactions with amounts bigger than 200 \euro{}. The anomaly score raised is the inverse of the likelihood of the recent amounts to have been generated by an HMM trained on a set of genuine sequences. They generated a credit card transactions dataset in order to assess the qualities of their strategies. \cite{dhok2012} used a similar approach for a real world credit card transaction dataset, and \cite{robinson2018} extended that with success for a prepaid credit card fraud detection task. 
\newline

\cite{dorj2013} also used HMMs in an anomaly detection task in electronic systems, but they modeled an abnormal sequence in addition to a genuine sequence. This is interesting since assessing the abnormality of outliers detected decreases the number of false positives. Not only do they flag outliers as such, but they also confirm that these outliers belong to the abnormal class. 
\newline

\cite{lucas2019ter} proposed a protocol for creating sequential features describing temporal dependencies between the transactions in a supervised way. For this purpose they isolated 3 different perspectives. First, they compare the likelihood of sequence of transactions against historical honest sequences of transactions but also against historical sequences containing at least one fraudulent transaction. The reasoning is that to have a risk of fraud it is not enough for a sequence to be far from honest behavior but it is also expected to be relatively close to a risky behavior. Second, they create these features in order to describe the card-holders sequences but also the merchants sequences. Credit card transactions can be seen as edges in a bipartite graph of merchants and card-holders and they showed that the efficiency of detection was better when taking into account both card-holder and terminal sequences. Third, they considered the elapsed time between two transactions and the amount of a transaction as signal for building their features.
Combinations of the three binary perspectives give eight sets of sequences from the (training) set of transactions. An Hidden Markov Model is trained on each one of these sequences. The proposed features is the likelihood that each the sequence of recent transactions have been generated by each HMM. They have shown a 15\% increase in detection compared to the transactions aggregation strategy of \cite{whitrow2008}.

\subsubsection{Recurrent neural networks}

Recurrent neural networks (RNNs) are a class of artificial neural networks where connections between nodes form a directed graph along a temporal sequence. These connections allow to model temporal dynamic behaviors.
\newline

Recurrent neural networks  can be considered as an hidden state based model where the inner layers reproduce the hidden state based architectures. Furthermore, the connectivity between consecutive hidden states (or nodes) also follows the Markov property ($p(y_{t})$ only depends on $y_{t-1}$). RNNs are discriminant models that aims to predict the label of an event given a past sequence of events. They are detailed extensively in \cite{graves2012}.
\newline

HMMs for sequential anomaly detection have mostly been used in order to raise an anomaly score using the likelihood of the sequence to have been generated by the HMM. However, recurrent neural network (RNN) approaches and more precisely long short term memory networks were more varied. Similarly to the previously mentioned HMM approaches, LSTM were used in order to raise anomaly scores by \cite{bontemps2016} for an intrusion detection task. The anomaly scores of a window of events is the distance between this window of events and the window of events that would have been generated by the LSTM for the corresponding user. Since LSTM are discriminative models, a custom distance has been created by the authors for this generative task. \cite{malhotra2015} modeled the genuine distribution of sequence based dataset with LSTM in order to afterwards raise an anomaly score for electrocardiograms, space shuttle valve time series and power demand datasets. Alternatively, \cite{ergen2017} used LSTM to extend various length sequences in order to obtain sequences of equal length for raising an anomaly score afterwards with the help of One Class SVM.

\cite{wiese2009} and \cite{jurgovsky2018} modeled the sequence of transactions in a card holder's account with Long short-term memory networks. In contrast to generative sequential models, the LSTM-based approaches are trained discriminatively by jointly learning a map from the sequence of transactions to a sequence of latent states and a map from latent states to the binary output variables (i.e. the fraud label of a transaction). The applications of LSTMs to credit card fraud detection are based on the assumption that the succession patterns of transactions bear relevant information for detecting frauds among them. Each latent state is considered a "summary" of the sequence of previously issued transactions and thereby it might permit a better classification of a recent purchase in the light of its summarized purchase sequence. However, in their comparison with order-agnostic feature aggregations, Jurgovsky et al.\cite{jurgovsky2018} found that feature aggregations yielded the best results overall - even in the card-present scenario in which consistent purchase patterns are more prevalent. 

LSTM were also used in order to raise anomaly scores by \cite{bontemps2016} for an intrusion detection task. The anomaly scores of a window of events is the distance between this window of events and the window of events that would have been generated by the LSTM for the corresponding user. Since LSTM are discriminative models, a custom distance has been created by the authors for this generative task. \cite{malhotra2015} modeled the genuine distribution of sequence based dataset with LSTM in order to afterwards raise an anomaly score for electrocardiograms, space shuttle valve time series and power demand datasets. Alternatively, \cite{ergen2017} used LSTM to extend various length sequences in order to obtain sequences of equal length for raising an anomaly score afterwards with the help of One Class SVM.

\subsection{Counting-based approaches}

Sequential anomaly detection aims to raise an anomaly score for events, subsequence of events or whole sequence with respect to a genuine distribution.
\newline

For sequential tasks, multiple definitions of anomalies exist that need to be handled differently (\cite{chandola2012}):
\begin{itemize}
\item An event of a sequence may be anomalous. For example, when a transaction is fraudulent within the sequence of transactions of a given card-holder.
\item A subsequence may be anomalous. For example, if there is a delay between the credit card theft and its reporting, the transactions issued with the credit card are not made by the card-holder but by the fraudster and must be identified as anomalous.
\item A sequence may be anomalous with respect to a set of non anomalous sequences. For example, a sentence in french will be anomalous in a set of english sentences.
\end{itemize}

One simple way to raise an anomaly score is counting the likelihood of the event or fixed length window of events in the genuine set of events. 
\begin{equation}
L_{i} = \frac{\# subsequence(i)}{\# subsequences}
\label{eq:oddratio}
\end{equation}

\cite{forrest1999} proposed the t-STIDE technique (threshold based sequence time delay embedding) in order to raise anomaly scores for sequences. It consists in computing the ratio of sliding windows that have a likelihood (see equation \ref{eq:oddratio}) below a certain threshold $\lambda$. The anomaly score of the test sequence is proportional to the number of anomalous windows in the test sequence. The corresponding formula is:
\begin{equation}
A_{Sq} = \frac{|i: L(w_{i})< \lambda , 1 \le i \le t |}{t} 
\end{equation} 

This kind of window based counting can also be used as anomaly signal to calculate a 2nd order anomaly score. For example, \cite{hofmeyr1998} proposed to compute the average of the strength of the anomaly signal over a set of windows that cover the considered event or subsequence.
\begin{equation}
A(Sq) = \frac{1}{N} \sum \limits_{i=1}^N A(w_{i})
\end{equation}

\section{Conclusion} \label{sec:conclusion}

Credit card fraud detection presents several intrinsic challenges. (1) There is a strong difference in terms of absolute numbers between the positive and the negative class: fraudulent transactions are usually much more rare than genuine transactions. (2) Purchase behaviours and fraudster strategies may change over time, making a learnt fraud detection decision function irrelevant if not updated. This phenomenon named dataset shift (change in the distribution $p( x, y )$ ) may prevent fraud detection systems from obtaining good performances. (3) The feature set describing a credit card transaction usually ignores detailed sequential information which was proven to be very relevant for the detection of credit card fraudulent transactions.

Different approaches for tackling each of these challenges are highlighted in this survey and for each of these approaches at minimum one research work is described in full detail. The goal is to provide the reader with useful information on all the different research subjects introduced here.

\bibliographystyle{plainnat}
\bibliography{literature.bib}

\begin{thebibliography}{91}
\providecommand{\natexlab}[1]{#1}
\providecommand{\url}[1]{\texttt{#1}}
\expandafter\ifx\csname urlstyle\endcsname\relax
  \providecommand{\doi}[1]{doi: #1}\else
  \providecommand{\doi}{doi: \begingroup \urlstyle{rm}\Url}\fi

\bibitem[Abdallah et~al.(2016)Abdallah, Maarof, and Zainal]{abdallah2016}
Aisha Abdallah, Mohd~Aizaini Maarof, and Anazida Zainal.
\newblock Fraud detection system: A survey.
\newblock \emph{Journal of Network and Computer Applications}, 2016.

\bibitem[Alaiz-Rodriguez and Japkowicz(2008)]{rodriguez2008}
Rocio Alaiz-Rodriguez and Nathalie Japkowicz.
\newblock Assessing the impact of changing environments on classifier
  performance.
\newblock \emph{Conference of the Canadian Society for Computational Studies of
  Intelligence, Springer}, 2008.

\bibitem[Ali et~al.(2019)Ali, Azad, Centeno, Hao, and van Moorsel]{Ali2019}
Mohammed~Aamir Ali, Muhammad~Ajmal Azad, Mario~Parreno Centeno, Feng Hao, and
  Aad van Moorsel.
\newblock Consumer-facing technology fraud: Economics, attack methods and
  potential solutions.
\newblock \emph{Future Generation Computer Systems (volume 100)}, 2019.

\bibitem[Bahnsen et~al.(2013)Bahnsen, Stojanovic, Aouada, and
  Ottersten]{bahnsen2013}
Alejandro~Correa Bahnsen, Aleksandar Stojanovic, Djamila Aouada, and
  Bj{\"{o}}rn Ottersten.
\newblock {Cost sensitive credit card fraud detection using bayes minimum
  risk}.
\newblock In \emph{Proceedings of the 12th International Conference on Machine
  Learning and Applications (ICMLA)}, volume~1, pages 333--338, 2013.
\newblock \doi{10.1109/ICMLA.2013.68}.

\bibitem[Bahnsen et~al.(2016)Bahnsen, Ouada, Stojanovic, and
  Ottersten]{bahnsen2016}
Alejandro~Correa Bahnsen, Djamila Ouada, Aleksandar Stojanovic, and Björn
  Ottersten.
\newblock Feature engineering strategies for credit card fraud detection.
\newblock \emph{Expert Systems with Applications}, 2016.

\bibitem[Barddal and Enembreck(2019)]{barddal2019bis}
Jean~Paul Barddal and Fabricio Enembreck.
\newblock Learning regularized hoeffding trees from data streams.
\newblock \emph{34th ACM/SIGAPP Symposium on Applied Computing (SAC2019)},
  2019.

\bibitem[Batista et~al.(2000)Batista, Carvalho, and Monard]{batista2000}
Gustavo Batista, Andre~Carlos Carvalho, and Maria~Carolina Monard.
\newblock Applying one-sided selection to unbalanced datasets.
\newblock \emph{MICAI 2000: Advances in Artificial Intelligence}, 2000.

\bibitem[Batista et~al.(2004)Batista, Prati, and Monard]{batista2004}
Gustavo Batista, Ronaldo Prati, and Maria~Carolina Monard.
\newblock A study of the behavior of several methods for balancing machine
  learning training data.
\newblock \emph{ACM SIGKDD Explorations Newsletter}, 2004.

\bibitem[Bhattacharyya et~al.(2011)Bhattacharyya, Jha, Tharakunnel, and
  Westland]{bhattacharyya2011}
Siddhartha Bhattacharyya, Sanjeev Jha, Kurian Tharakunnel, and J.~Christopher
  Westland.
\newblock Data mining for credit card fraud: A comparative study.
\newblock \emph{Decision support systems}, 2011.

\bibitem[Bickel et~al.(2007)Bickel, Brückner, and Scheffer]{bickel2007}
Steffen Bickel, Michael Brückner, and Tobias Scheffer.
\newblock Discriminative learning for differing training and test
  distributions.
\newblock \emph{Proceedings of the 24th International conference on Machine
  learning}, 2007.

\bibitem[Bifet and Gavalda(2009)]{bifet2009}
A.~Bifet and R.~Gavalda.
\newblock Adaptive learning from evolving data streams.
\newblock \emph{International Symposium on Intelligent Data Analysis}, 2009.

\bibitem[Bishop(2006)]{bishop2006pattern}
Christopher Bishop.
\newblock \emph{{Pattern Recognition and Machine Learning}}.
\newblock Springer, New York, 1 edition, 2006.
\newblock ISBN 978-0-387-31073-2.

\bibitem[Blei et~al.(2001)Blei, Ng, and Jordan]{blei2003latent}
David~M. Blei, Andrew~Y. Ng, and Michael~I. Jordan.
\newblock {Latent Dirichlet Allocation}.
\newblock \emph{Proceedings of the 14th Annual Conference on Neural Information
  Processing Systems}, 3\penalty0 (Jan):\penalty0 601--608, 2001.
\newblock ISSN 1532-4435.
\newblock \doi{10.1162/jmlr.2003.3.4-5.993}.

\bibitem[Bolton and Hand(2001)]{bolton2001}
Richard~J. Bolton and David~J. Hand.
\newblock Unsupervised profiling methods for fraud detection.
\newblock \emph{Credit Scoring and Credit Control}, 2001.

\bibitem[Bontemps et~al.(2016)Bontemps, Cao, McDermott, and
  Le-Khac]{bontemps2016}
Loic Bontemps, Van~Loi Cao, James McDermott, and Nhien-An Le-Khac.
\newblock Collective anomaly detection based on long short-term memory
  recurrent neural networks.
\newblock \emph{International Conference on Future Data and Security
  Engineering}, 2016.

\bibitem[Chandola et~al.(2012)Chandola, Banerjee, and Kumar]{chandola2012}
Varun Chandola, Arindam Banerjee, and Vipin Kumar.
\newblock Anomaly detection for discrete sequences: A survey.
\newblock \emph{IEEE Transactions on Knowledge and Data Engineering}, 2012.

\bibitem[Chawla(2005)]{chawla2005}
Nitesh~V. Chawla.
\newblock Data mining for imbalance datasets: an overview.
\newblock \emph{Data mining and knowledge discovery handbook}, 2005.

\bibitem[Chawla et~al.(2002)Chawla, Bowyer, Hall, and Kegelmeyer]{chawla2002}
Nitesh~V. Chawla, Kevin~W. Bowyer, Lawrence~O. Hall, and W.~Philip Kegelmeyer.
\newblock Smote: Synthetic minority over-sampling technique.
\newblock \emph{Journal of Artificial Intelligence Research 16}, 2002.

\bibitem[Chawla et~al.(2003)Chawla, Lazarevic, Bowyer, and Hall]{chawla2003}
Nitesh~V. Chawla, Aleksandar Lazarevic, Kevin~W. Bowyer, and Lawrence~O. Hall.
\newblock Smoteboost: Improving prediction of the minority class in boosting.
\newblock \emph{Knowledge Discovery in Databases: PKDD}, 2003.

\bibitem[Chawla et~al.(2004)Chawla, Japkowicz, and Kotcz]{chawla2004}
Nitesh~V. Chawla, Nathalie Japkowicz, and Aleksander Kotcz.
\newblock Editorial: special issue on learning from imbalanced data sets.
\newblock \emph{ACM SIGKDD Explorations Newsletter}, 2004.

\bibitem[Davis and Goadrich(2006)]{Davis2006}
Jesse Davis and Mark Goadrich.
\newblock The relationship between precision-recall and roc curves.
\newblock \emph{ICML '06 Proceedings of the 23rd international conference on
  Machine learning}, 2006.

\bibitem[Dhok(2012)]{dhok2012}
Shailesh~S. Dhok.
\newblock Credit card fraud detection using hidden markov model.
\newblock \emph{International Journal of Soft Computing and Engineering}, 2012.

\bibitem[Dorj et~al.(2013)Dorj, Chen, and Pecht]{dorj2013}
Enkhjargal Dorj, Chaochao Chen, and Michael Pecht.
\newblock A bayesian hidden markov model-based approach for anomaly detection
  in electronic systems.
\newblock \emph{IEEE Aerospace conference}, 2013.

\bibitem[Drummond and Holte(2003)]{drummond2003}
Chris Drummond and Robert Holte.
\newblock C4.5, class imbalance, and cost sensitivity: why undersampling beats
  over-sampling.
\newblock \emph{Workshop on Learning from Imbalanced Datasets II}, 2003.

\bibitem[Elkan(2001)]{elkan2001}
Charles Elkan.
\newblock The foundations of cost-sensitive learning.
\newblock \emph{International Joint Conference on Artificial Intelligence},
  2001.

\bibitem[Ergen et~al.(2017)Ergen, Mirza, and Kozat]{ergen2017}
Tolga Ergen, Ali~Hassan Mirza, and Suleyman~Serdar Kozat.
\newblock Unsupervised and semi-supervised anomaly detection with lstm neural
  networks.
\newblock \emph{arXiv preprint}, 2017.

\bibitem[Estabrooks et~al.(2004)Estabrooks, Jo, and Japkowicz]{estabrooks2004}
Andrew Estabrooks, Taeho Jo, and Nathalie Japkowicz.
\newblock A multiple resampling method for learning from imbalanced data sets.
\newblock \emph{Computational Intelligence}, 2004.

\bibitem[Fan et~al.(1999)Fan, Stolfo, Zhang, and Chan]{wei1999}
Wei Fan, Salvatore~J Stolfo, Junxin Zhang, and Philip~K Chan.
\newblock Adacost: misclassification cost-sensitive boosting.
\newblock \emph{ICML}, 1999.

\bibitem[Forrest et~al.(1999)Forrest, Warrender, and Pearlmutter]{forrest1999}
S.~Forrest, C.~Warrender, and B.~Pearlmutter.
\newblock Detecting intrusions using system calls: Alternate data models.
\newblock \emph{Proceedings of the 1999 IEEE ISRSP}, 1999.

\bibitem[Fu et~al.(2016)Fu, Cheng, Tu, and Zhang]{Fu2016}
Kang Fu, Dawei Cheng, Yi~Tu, and Liqing Zhang.
\newblock Credit card fraud detection using convolutional neural networks.
\newblock In Akira Hirose, Seiichi Ozawa, Kenji Doya, Kazushi Ikeda, Minho Lee,
  and Derong Liu, editors, \emph{Neural Information Processing}, pages
  483--490, Cham, 2016. Springer International Publishing.
\newblock ISBN 978-3-319-46675-0.

\bibitem[Gao et~al.(2007)Gao, Fan, Han, and Yu]{gao2007}
J~Gao, W~Fan, J~Han, and PS~Yu.
\newblock A general framework for mining concept-drifting data streams with
  skewed distributions.
\newblock \emph{Proceedings of the 2007 SIAM International Conference}, 2007.

\bibitem[Gomes et~al.(2017)Gomes, Bifet, Read, Barddal, Enembreck, Pfharinger,
  Holmes, and Abdessalem]{gomes2017}
Heitor~M. Gomes, Albert Bifet, Jesse Read, Jean~Paul Barddal, Fabricio
  Enembreck, Bernhard Pfharinger, Geoff Holmes, and Talel Abdessalem.
\newblock Adaptive random forest for evolving data stream classification.
\newblock \emph{Machine Learning 106(9)}, 2017.

\bibitem[Graves(2012)]{graves2012}
A.~Graves.
\newblock Supervised sequence labelling with recurrent neural networks.
\newblock \emph{Springer vol. 385}, 2012.

\bibitem[Gretton et~al.(2009)Gretton, Smola, Huang, Schmittful, and
  Borgwardt]{gretton2009}
Arthur Gretton, Alex Smola, Jiayuan Huang, M~Schmittful, and K~Borgwardt.
\newblock Covariate shift by kernel mean matching.
\newblock \emph{Dataset shift in machine learning 3(4)}, 2009.

\bibitem[Han et~al.(2005)Han, Wang, and Mao]{han2005}
Hui Han, Wen-Yuan Wang, and Bing-Huan Mao.
\newblock Borderline-smote: a new oversampling method in imbalanced data sets
  learning.
\newblock \emph{Advances in intelligent computing, Springer}, 2005.

\bibitem[Hand(2009)]{Hand2009}
David~J. Hand.
\newblock {Measuring classifier performance: A coherent alternative to the area
  under the ROC curve}.
\newblock \emph{Machine Learning}, 77\penalty0 (1):\penalty0 103--123, 2009.
\newblock ISSN 08856125.
\newblock \doi{10.1007/s10994-009-5119-5}.
\newblock URL \url{https://doi.org/10.1007/s10994-009-5119-5}.

\bibitem[He et~al.(2008)He, Bai, Garcia, and Li]{he2008}
Haibo He, Yang Bai, Edwardo Garcia, and Shutao Li.
\newblock Adasyn: Adaptive synthetic sampling approach for imbalanced learning.
\newblock \emph{IEEE International Joint Conference on Neural Networks}, 2008.

\bibitem[Hoens et~al.(2011)Hoens, Chawla, and Polikar]{hoens2011}
Ryan~T. Hoens, Nitesh~V. Chawla, and Robi Polikar.
\newblock Heuristic updatable weighted random subspaces for non-stationary
  environments.
\newblock \emph{IEEE 11th International Conference on Data Mining}, 2011.

\bibitem[Hofmeyr et~al.(1998)Hofmeyr, Forrest, and Somayaji]{hofmeyr1998}
S.~A. Hofmeyr, S.~Forrest, and A.~Somayaji.
\newblock Intrusion detection using sequences of system calls.
\newblock \emph{Journal of Computer Security}, 1998.

\bibitem[Japkowicz and Stephen(2002)]{japkowicz2002}
Nathalie Japkowicz and Shaju Stephen.
\newblock The class imbalance problem: a systematic study.
\newblock \emph{Intelligent data analysis}, 2002.

\bibitem[Jha et~al.(2012)Jha, Guillen, and Westland]{jha2012}
S.~Jha, M.~Guillen, and J.~Westland.
\newblock Employing transaction aggregatin strategy to detect credit card
  fraud.
\newblock \emph{Expert Systems with Applications}, 2012.

\bibitem[Jones(1973)]{JONES1973619}
Karen~Sparck Jones.
\newblock {Index term weighting}.
\newblock \emph{Information Storage and Retrieval}, 9\penalty0 (11):\penalty0
  619--633, 1973.
\newblock ISSN 00200271.
\newblock \doi{10.1016/0020-0271(73)90043-0}.
\newblock URL
  \url{http://www.sciencedirect.com/science/article/pii/0020027173900430}.

\bibitem[Jurgovsky(2020)]{jurgovskythesis}
Johannes Jurgovsky.
\newblock Context aware credit card fraud detection.
\newblock \emph{Thesis dissertation}, 2020.

\bibitem[Jurgovsky et~al.(2018)Jurgovsky, Granitzer, Ziegler, Calabretto,
  Portier, He-Guelton, and Caelen]{jurgovsky2018}
Johannes Jurgovsky, Michael Granitzer, Konstantin Ziegler, Sylvie Calabretto,
  Pierre-Edouard Portier, Liyun He-Guelton, and Olivier Caelen.
\newblock Sequence classification for credit-card fraud detection.
\newblock \emph{Expert systems with applications}, 2018.

\bibitem[Karax et~al.(2019)Karax, Malucelli, and Barddal]{barddal2019}
Jean Antonio~Pereira Karax, Andreia Malucelli, and Jean~Paul Barddal.
\newblock Decision tree-based feature ranking in concept drifting data streams.
\newblock \emph{34th ACM/SIGAPP Symposium on Applied Computing (SAC2019)},
  2019.

\bibitem[Kelly et~al.(1999)Kelly, Hand, and Adams]{kelly1999}
Mark~G. Kelly, David~J. Hand, and Niail~M. Adams.
\newblock The impact of changing populations on classifier performance.
\newblock \emph{Proceedings of the fifth ACM SIGKDD international conference on
  Knowledge discovery and data mining ACM}, 1999.

\bibitem[Klinkenberg(2003)]{klinkenberg2004}
Ralf Klinkenberg.
\newblock Learning drifting concepts: Example selection vs. example weighting.
\newblock \emph{Intelligent Data Analysis vol 8}, 2003.

\bibitem[Kolter and Maloof(2003)]{kolter2003}
Jeremy~Z. Kolter and Marcus~A. Maloof.
\newblock Dynamic weighted majority: A new ensemble method for tracking concept
  drift.
\newblock \emph{Third IEEE International conference on data mining}, 2003.

\bibitem[Kotsiantis(2007)]{kotsiantis2007}
Sotiris Kotsiantis.
\newblock Supervised machine learning: A review of classification techniques.
\newblock \emph{Informatica}, 2007.

\bibitem[Krivko(2010)]{krivko2010hybrid}
M.~Krivko.
\newblock {A hybrid model for plastic card fraud detection systems}.
\newblock \emph{Expert Systems with Applications}, 37\penalty0 (8):\penalty0
  6070--6076, 2010.
\newblock ISSN 09574174.
\newblock \doi{10.1016/j.eswa.2010.02.119}.

\bibitem[Kubat and Matwin(1997)]{kubat1997}
Miroslav Kubat and Stan Matwin.
\newblock Addressing the curse of imbalanced training sets: one-sided
  selection.
\newblock \emph{ICML}, 1997.

\bibitem[Kukar and Kononenko(1998)]{kukar1998}
Matjaz Kukar and Igor Kononenko.
\newblock Cost-sensitive learning with neural networks.
\newblock \emph{ECAI}, 1998.

\bibitem[Lam et~al.(2015)Lam, Wei, and Wunsch]{lam2015clustering}
Dao Lam, Mingzhen Wei, and Donald Wunsch.
\newblock {Clustering Data of Mixed Categorical and Numerical Type With
  Unsupervised Feature Learning}.
\newblock \emph{IEEE Access}, 3:\penalty0 1605--1613, 2015.
\newblock \doi{10.1109/ACCESS.2015.2477216}.

\bibitem[Lane and Brodley(1998)]{lane1998}
Terran Lane and Carla~E. Brodley.
\newblock Approaches to online learing and concept drift for user
  identification in computer security.
\newblock \emph{KDD}, 1998.

\bibitem[Lichtenwalter and Chawla(2009)]{lichtenwalter2009}
Ryan~N. Lichtenwalter and Nitesh~V. Chawla.
\newblock Adaptive methods for classification in arbitrarily imbalanced and
  drifting data streams.
\newblock \emph{Pacific-Asia Conference on Knowledge Discovery and Data
  Mining}, 2009.

\bibitem[Liu et~al.(2009)Liu, Wu, and Zhou]{liu2009}
Xu-Ying Liu, Jianxin Wu, and Zhi-Hua Zhou.
\newblock Exploratory undersampling for class-imbalance learning.
\newblock \emph{Systems, Man, and Cybernetics, part B: Cybernetics}, 2009.

\bibitem[Lucas(2020)]{Lucas2020}
Yvan Lucas.
\newblock Credit card fraud detection using machine learning with integration
  of contextual knowledge.
\newblock \emph{PhD Thesis}, 2020.

\bibitem[Lucas et~al.(2019)Lucas, Portier, Laporte, Calabretto, Caelen,
  He-Guelton, and Granitzer]{lucas2019}
Yvan Lucas, Pierre-Edouard Portier, Léa Laporte, Sylvie Calabretto, Olivier
  Caelen, Liyun He-Guelton, and Michael Granitzer.
\newblock Multiple perspectives hmm-based feature engineering for credit card
  fraud detection.
\newblock \emph{34th ACM/SIGAPP Symposium on Applied Computing (SAC2019)},
  2019.

\bibitem[Lucas et~al.(2020)Lucas, Portier, Laporte, Calabretto, Caelen,
  He-Guelton, and Granitzer]{lucas2019ter}
Yvan Lucas, Pierre-Edouard Portier, Léa Laporte, Sylvie Calabretto, Olivier
  Caelen, Liyun He-Guelton, and Michael Granitzer.
\newblock Towards automated feature engineering for credit card fraud detection
  using multi-perspective hmms.
\newblock \emph{Future Generations Computer Systems Special Issue on: Data
  Exploration in the Web 3.0 Age}, 2020.

\bibitem[Malhotra et~al.(2015)Malhotra, Vig, Shroff, and Agarwal]{malhotra2015}
Pankaj Malhotra, Lovekesh Vig, Gautam Shroff, and Puneet Agarwal.
\newblock Long short term memory networks for anomaly detection in time series.
\newblock \emph{European Symposium on Artificial Neural Networks, Computational
  Intelligence and Machine Learning}, 2015.

\bibitem[Mani and Zhang(2003)]{mani2003}
Inderjeet Mani and I~Zhang.
\newblock knn approach to unbalanced data distributions: a case study involving
  information extraction.
\newblock \emph{Proceedings of Workshop on Learning from Imbalanced Datasets},
  2003.

\bibitem[Mikolov et~al.(2011)Mikolov, Deoras, Kombrink, Burget, and
  Cernocky]{mikolov2011}
Tomas Mikolov, Anoop Deoras, Stefan Kombrink, Lukas Burget, and Jan Cernocky.
\newblock Empirical evaluation and combination of advanced language modeling
  techniques.
\newblock \emph{12th Annual Conference of the International Speech
  Communication Association}, 2011.

\bibitem[Moreno-Torres et~al.(2012)Moreno-Torres, Raeder, Alaiz-Rodriguez,
  Chawla, and Herrera]{torres2012}
JG~Moreno-Torres, T~Raeder, R~Alaiz-Rodriguez, N~V Chawla, and F~Herrera.
\newblock A unifying view on dataset shift in classification.
\newblock \emph{Pattern recognition}, 2012.

\bibitem[Moreno-Torres et~al.(2013)Moreno-Torres, Llora, Goldberg, and
  Bhargava]{torres2013}
Jose~G. Moreno-Torres, Xavier Llora, David~E. Goldberg, and Rohit Bhargava.
\newblock Repairing fractures between data using genetic programming-based
  feature extraction: A case study in cancer diagnosis.
\newblock \emph{Information Sciences 222}, 2013.

\bibitem[Pastor and Baralis(2019)]{pastor2019}
Eliana Pastor and Elena Baralis.
\newblock Explaining black box models by means of local rules.
\newblock \emph{34th ACM/SIGAPP Symposium on Applied Computing (SAC2019)},
  2019.

\bibitem[Power(2011)]{david2011}
David Power.
\newblock Evaluation: from precision, recall and f-measure to roc,
  informedness, markedness and correlation.
\newblock \emph{Journal of machine learning technology (1)}, 2011.

\bibitem[Pozzolo et~al.(2014)Pozzolo, Caelen, and Borgne]{pozzolo2014}
A~Dal Pozzolo, O~Caelen, and YA~Le Borgne.
\newblock Learned lessons in credit card fraud detection from a practitioner
  perspective.
\newblock \emph{Expert systems with applications}, 2014.

\bibitem[Pozzolo(2015)]{pozzolo2015}
Andrea~Dal Pozzolo.
\newblock Adaptive machine learning for credit card fraud detection.
\newblock \emph{PhD Thesis}, 2015.

\bibitem[Pozzolo et~al.(2015)Pozzolo, Caelen, Johnson, and
  Bontempi]{dal2015calibrating}
Andrea~Dal Pozzolo, Olivier Caelen, Reid~A. Johnson, and Gianluca Bontempi.
\newblock {Calibrating probability with undersampling for unbalanced
  classification}.
\newblock In \emph{In Proceedings of the IEEE Symposium Series on Computational
  Intelligence (SSCI)}, pages 159--166. IEEE, 2015.
\newblock ISBN 9781479975600.
\newblock \doi{10.1109/SSCI.2015.33}.

\bibitem[Robinson and Aria(2018)]{robinson2018}
William~N. Robinson and Andrea Aria.
\newblock Sequential fraud detection for prepaid cards using hidden markov
  model divergence.
\newblock \emph{Expert Systems with Applications}, 2018.

\bibitem[Russac et~al.(2018)Russac, Caelen, and He-Guelton]{russac2018}
Yoan Russac, Olivier Caelen, and Liyun He-Guelton.
\newblock Embeddings of categorical variables for sequential data in fraud
  context.
\newblock \emph{International Conference on Advanced Machine Learning
  Technologies and Applications}, 2018.

\bibitem[Sahin et~al.(2013)Sahin, Bulkan, and Duman]{sahin2013cost}
Yusuf Sahin, Serol Bulkan, and Ekrem Duman.
\newblock {A cost-sensitive decision tree approach for fraud detection}.
\newblock \emph{Expert Systems with Applications}, 40\penalty0 (15):\penalty0
  5916--5923, 2013.
\newblock ISSN 09574174.
\newblock \doi{10.1016/j.eswa.2013.05.021}.
\newblock URL \url{http://dx.doi.org/10.1016/j.eswa.2013.05.021}.

\bibitem[Saito and Rehmsmeier(2015)]{NoROCSaito2015}
Takaya Saito and Marc Rehmsmeier.
\newblock {The Precision-Recall Plot Is More Informative than the ROC Plot When
  Evaluating Binary Classifiers on Imbalanced Datasets}.
\newblock \emph{PLOS ONE}, 10\penalty0 (3):\penalty0 1--21, 2015.
\newblock \doi{10.1371/journal.pone.0118432}.
\newblock URL \url{https://doi.org/10.1371/journal.pone.0118432}.

\bibitem[Shimodaira(2000)]{shimodaira2000}
Hidetoshi Shimodaira.
\newblock Improving predictive inference under covariate shift by weighting the
  log-likelihood function.
\newblock \emph{Journal of statistical planning and inference 90(2)}, 2000.

\bibitem[Shirazi and Vasconcelos(2010)]{shirazi2010}
Hamed~Masnadi Shirazi and Nuno Vasconcelos.
\newblock Risk minimization, probability elicitation and cost-sensitive svms.
\newblock \emph{ICML}, 2010.

\bibitem[Srivastava et~al.(2008)Srivastava, Kundu, Sural, and
  Majumdar]{srivastava2008}
Abhinav Srivastava, Amlan Kundu, Shamik Sural, and Arun~K. Majumdar.
\newblock Credit card fraud detection using hidden markov model.
\newblock \emph{IEEE Transactions on dependable and secure computing}, 2008.

\bibitem[Storkey(2009)]{storkey2009}
Amos Storkey.
\newblock When training and test sets are different: characterizing learning
  transfer.
\newblock \emph{Dataset shift in machine learning}, 2009.

\bibitem[Sugiyama et~al.(2007)Sugiyama, Krauledat, and Mazller]{sugiyama2007}
Masashi Sugiyama, Matthias Krauledat, and Klaus-Robert Mazller.
\newblock Covariate shift adaptation by importance weighted cross validation.
\newblock \emph{Journal of Machine Learning Research 8}, 2007.

\bibitem[Sun et~al.(2007)Sun, Kamel, Wong, and Wang]{sun2007}
Yanmin Sun, Mohamed~S Kamel, Andrew~KC Wong, and Yang Wang.
\newblock Cost-sensitive boosting for classification of imbalanced data.
\newblock \emph{Pattern Recognition}, 2007.

\bibitem[Visa and Ralescu(2005)]{visa2005}
Sofia Visa and Anca Ralescu.
\newblock Issues in mining imbalanced data sets: a review paper.
\newblock \emph{Proceedings of the sixteen midwest artificial intelligence and
  cognitive science conference}, 2005.

\bibitem[Vlasselaer et~al.(2015)Vlasselaer, Bravo, Caelen, Eliassirad, Akoglu,
  Snoeck, and Baesens]{vlasselaer2015}
Veronique~Van Vlasselaer, Cristian Bravo, Olivier Caelen, TIna Eliassirad,
  Leman Akoglu, Monique Snoeck, and Bart Baesens.
\newblock Apate: A novel approach for automated credit card transactions fraud
  detection using network-based extensions.
\newblock \emph{Decision support systems}, 2015.

\bibitem[Wang et~al.(2003)Wang, Zhou, Fu, and Yu]{wang2003}
Ke~Wang, Senqiang Zhou, Chee~Ada Fu, and J.X. Yu.
\newblock Mining changes of classification by correspondance tracing.
\newblock \emph{Proceedings of the 2003 SIAM International COnference on Data
  Mining. Society for Industrial and Applied Mathematics}, 2003.

\bibitem[Wang et~al.(2009)Wang, Tang, and Yao]{wang2009}
Shuo Wang, Ke~Tang, and Xin Yao.
\newblock Diversity exploration and negative correlation learning on imbalanced
  data sets.
\newblock \emph{International Joint Conference on Neural Networks}, 2009.

\bibitem[Webb and Ting(2005)]{webb2005}
Geoffrey~I. Webb and Kai~Ming Ting.
\newblock On the application of roc analysis to predict classification
  performance under varying class distributions.
\newblock \emph{Machine Learning 58}, 2005.

\bibitem[Weiss and Provost(2001)]{weiss2001}
Gary~M Weiss and Foster Provost.
\newblock The effect of class distribution on classifier learning: an empirical
  study.
\newblock \emph{Rutgers Univ}, 2001.

\bibitem[Whitrow et~al.(2008)Whitrow, Hand, Juszczak, Weston, and
  Adams]{whitrow2008}
C.~Whitrow, D.~J. Hand, P.~Juszczak, D.~J. Weston, and N.~M. Adams.
\newblock Transaction aggregation as a strategy for credit card fraud
  detection.
\newblock \emph{Data Mining and Knowledge Discovery 18(1)}, 2008.

\bibitem[Widmer and Kubat(1996)]{widmer1996}
Gerhard Widmer and Miroslav Kubat.
\newblock Learning in the presence of context drift and hidden contexts.
\newblock \emph{Machine Learning 23}, 1996.

\bibitem[Wiese and Omlin(2009)]{wiese2009}
B.~Wiese and C.~Omlin.
\newblock Credit card transactions, fraud detection and machine learning:
  Modelling time with lstm recurrent neural networks.
\newblock \emph{Innovations in neural information paradigms and applications,
  Springer}, 2009.

\bibitem[Wu and Chang(2003)]{wu2003}
Gang Wu and Edward~Y Chang.
\newblock Class-boundary alignment for imbalanced dataset learning.
\newblock \emph{ICML workshop on learning from imbalanced data sets II}, 2003.

\bibitem[Yan et~al.(2003)Yan, Liu, Jin, and Hauptmann]{yan2003}
Rong Yan, Yan Liu, Rong Jin, and Alex Hauptmann.
\newblock On predicting rare classes with svm ensembles in scene
  classification.
\newblock \emph{IEEE International Conference on Acoustics, Speech, and Signal
  Processing}, 2003.

\bibitem[Ziegler et~al.(2017)Ziegler, Caelen, Garchery, Granitzer, He-Guelton,
  Jurgovsky, Portier, and Zwicklbauer]{wetice2017}
K~Ziegler, O~Caelen, M~Garchery, M~Granitzer, L~He-Guelton, J~Jurgovsky,
  P~Portier, and S~Zwicklbauer.
\newblock {Injecting Semantic Background Knowledge into Neural Networks using
  Graph Embeddings}.
\newblock In \emph{26th IEEE International Conference on Enabling Technologies:
  Infrastructure for Collaborative Enterprises. (WETICE).}, pages 200--205,
  2017.
\newblock \doi{10.1109/WETICE.2017.36}.

\end{thebibliography}

\end{document}